\documentclass{article}
\usepackage{amsmath,amsfonts,bm}

\def\eqref#1{equation~\ref{#1}}
\def\1{\bm{1}}

\def\va{{\bm{a}}}

\def\vh{{\bm{h}}}

\def\vm{{\bm{m}}}

\def\vx{{\bm{x}}}

\def\mW{{\bm{W}}}

\DeclareMathAlphabet{\mathsfit}{\encodingdefault}{\sfdefault}{m}{sl}
\SetMathAlphabet{\mathsfit}{bold}{\encodingdefault}{\sfdefault}{bx}{n}

\def\sR{{\mathbb{R}}}

\usepackage{microtype}
\usepackage{graphicx}
\usepackage{subfigure}
\usepackage{booktabs} % for professional tables

\usepackage{hyperref}
\usepackage{url}
\usepackage{multirow}
\usepackage[flushleft]{threeparttable}
\usepackage{longtable}
\usepackage{supertabular}
\usepackage{colortbl}
\usepackage{xspace}
\usepackage{wrapfig}
\usepackage{xcolor}
\usepackage{cancel}
\usepackage{amsthm}

\newtheorem{definition}{Definition}

\newcommand{\ie}{\textit{i}.\textit{e}.,}
\newcommand{\eg}{\textit{e}.\textit{g}.}
\newcommand{\ours}{\textsc{GraphArm}\xspace}
\usepackage{hyperref}

\usepackage[accepted]{icml2023}

\icmltitlerunning{Autoregressive Diffusion Model for Graph Generation}

\begin{document}

\twocolumn[
\icmltitle{Autoregressive Diffusion Model for Graph Generation}




\begin{icmlauthorlist}
\icmlauthor{Lingkai Kong}{ga}
\icmlauthor{Jiaming Cui}{ga}
\icmlauthor{Haotian Sun}{ga}
\icmlauthor{Yuchen Zhuang}{ga}
\icmlauthor{B. Aditya Prakash}{ga}
\icmlauthor{Chao Zhang}{ga}
\end{icmlauthorlist}

\icmlaffiliation{ga}{School of Computational Science and Engineering, Georgia Institute of Technology, Atlanta, USA}

\icmlcorrespondingauthor{Lingkai Kong}{lkkong@gatech.edu}

\icmlkeywords{Machine Learning, ICML}

\vskip 0.3in
]



\printAffiliationsAndNotice{} % otherwise use the standard text.

\begin{abstract}

  Diffusion-based graph generative models have recently obtained promising results for graph generation.
  However, existing diffusion-based graph generative models are mostly one-shot generative models that apply Gaussian diffusion in the  dequantized adjacency matrix space.
  Such a strategy can suffer from difficulty in model training, slow sampling speed, and incapability of incorporating constraints.  We propose an \emph{autoregressive diffusion} model for graph generation.
  Unlike existing methods, we define a node-absorbing diffusion process that operates directly in the discrete graph space.
  For forward diffusion,
  we design a \emph{diffusion ordering network}, which learns a data-dependent node  absorbing ordering from graph topology.
  For reverse generation, we design a \emph{denoising network} that uses the reverse node ordering to efficiently reconstruct the graph by  predicting the node type of the new node and its edges with previously denoised nodes at a time.
  Based on the permutation invariance of graph, we show that the two networks can be jointly trained by optimizing a simple lower bound of data likelihood. Our experiments on six diverse generic graph datasets and two molecule datasets show that our model achieves better or comparable generation performance with previous state-of-the-art, and meanwhile enjoys fast generation speed.

\end{abstract}

\section{Introduction}
\label{sec:intro}
Generating graphs from a target distribution is a fundamental problem in many
domains such as drug discovery \citep{li2018learning}, material design
\citep{maziarka2020mol}, social network analysis \citep{grover2019graphite},
and public health \citep{yu2020reverse}. Deep generative models have recently
led to promising advances in this problem. Different from traditional random
graph models \citep{erdos1960evolution, albert2002statistical}, these methods
fit graph data with powerful deep generative models including variational
auto-encoders (VAEs) \citep{simonovsky2018graphvae}, generative adversarial networks (GANs)
\citep{maziarka2020mol}, normalizing flows \citep{madhawa2019graphnvp}, and
energy-based models (EBMs) \citep{liu2021graphebm}. These models are learned to
capture complex graph structural patterns and then generate new high-fidelity
graphs with desired properties \cite{zhusurvey, du2021graphgt}.

Recently, the emergence of probabilistic diffusion models has led to interest
in diffusion-based graph generation \citep{jo2022score}. Diffusion models
decompose the full complex transformation between noise and real data into
many small steps of simple diffusion. Compared with prior deep generative
models, diffusion models enjoy both flexibility in modeling architecture and
tractability of the model's probability distributions.  However, existing diffusion-based graph generative models \cite{niu2020permutation,jo2022score, vignac2022digress} suffer from three
key drawbacks: (1) \emph{Generation Efficiency}. The sampling processes
 are slow, as they require a very long diffusion process to arrive at the stationary noisy distribution and thus the reverse generation process is also time-consuming.  (2) \emph{Incorporating constraints}. They are
all one-shot generation models and hence cannot easily incorporate
constraints during the one-shot generation process.
(3) \emph{Continuous Approximation.}
\citet{niu2020permutation, jo2022score} convert discrete graphs to continuous state spaces by adding real-valued
noise to graph adjacency matrices. Such dequantization can distort the
distribution of the original discrete graph structures, thus increasing the
difficulty of model training.

We propose an autoregressive graph generative model named \ours via
\emph{autoregressive diffusion} on graphs. Autoregressive diffusion  model (ARDM)
\citep{hoogeboom2022autoregressive} builds upon the recently developed absorbing diffusion
\citep{austin2021structured} for discrete data, where exactly one dimension of
the data decays to the absorbing state at each diffusion step. In \ours, we
design \emph{node-absorbing autoregressive diffusion} for graphs, which
diffuses a graph directly in the discrete graph space instead of in the
dequantized adjacency matrix space. The forward pass absorbs one node in each
step by masking it along with its connecting edges,
which is repeated until all the nodes are absorbed and the graph becomes
empty. We further design a \emph{diffusion ordering network} in \ours, which is jointly trained with the reverse generator to learn a data-dependent node ordering for  diffusion. Compared with random ordering
as in prior absorbing diffusion \citep{hoogeboom2022autoregressive}, the learned
diffusion ordering not only provides a better approximation
of the true marginal graph likelihood, but also eases the generative model
training by leveraging  structural regularities.
The backward pass in \ours recovers the graph structure by learning to reverse
the node-absorbing diffusion process with a denoising network. The reverse
generative process is autoregressive, which makes \ours easier to handle the constraints during generation. However, a key challenge is to learn the
distribution of reverse node ordering for optimizing the data likelihood. We
show that this difficulty can be circumvented by just using the exact reverse
node ordering and optimizing a simple lower bound of likelihood, based on the
permutation invariance property of graph generation. The likelihood lower
bound allows for jointly training the denoising network and the diffusion
ordering network using a reinforcement learning procedure and gradient
descent.

The generation speed of \ours is much faster than the existing graph diffusion
models \citep{jo2022score,niu2020permutation,vignac2022digress}. Due to the autoregressive
diffusion process in the node space, the number of diffusion steps in \ours is
the same as the number of nodes, which is typically much smaller than the
sampling steps in \citep{jo2022score,niu2020permutation,vignac2022digress}. Furthermore, at each
step of the backward pass, we design the denoising network to predict the node type of the newly generated node and its edges with previously denoised nodes at one time. The edges to be predicted follow a
mixture of multinomial distribution to ensure dependencies among each other.
This makes \ours offer a more balanced trade-off between flexibility and efficiency.

Our key contributions are as follows: (1) To the best of our knowledge, our
work is the first \emph{autoregressive} diffusion-based graph generation
model, underpinned by a new node self-absorbing diffusion process. Our model represents a generalized form of the diffusion processes, a class in which ARDM \cite{hoogeboom2022autoregressive} and diffusion Schrödinger bridge \cite{de2021diffusion} also fall. (2) \ours
learns a data-dependent node generation ordering and thus better leverages the
structural regularities for autoregressive graph diffusion. (3) We validate our
method on eight graph generation tasks, on which we
show that \ours outperforms existing graph generative models and is efficient
in generation speed.

\vspace{-2mm}
\section{Additional Related Work}
\vspace{-1mm}
% \subsection{Graph Generative Models}

% Existing deep graph generative models can be categorized into two classes. The
% first class is

% Though these models enjoy fast generation speed, they typically
% suffer from limited generation quality.

\paragraph{Graph Generation.}

\emph{One-shot graph generative models} generate all edges between nodes at
once. Models based on VAEs~\citep{simonovsky2018graphvae,
  liu2018constrained,maco2018} and GANs~\citep{caomo2022,maziarka2020mol}
generate all edges independently from latent embeddings. This independence
assumption can hurt the quality of the generated graphs. Normalizing flow
models \citep{zang2020moflow,madhawa2019graphnvp} are restricted to invertible
model architectures for building a normalized probability. The other class is \emph{autoregressive graph generative models}, which generate a graph by sequentially adding nodes and edges. Autoregressive generation can be achieved using recurrent networks \citep{li2018learning,you2018graphrnn, dai2020scalable},
VAEs~\citep{liu2018constrained,jinju2018,jinhi2020, guo2021deep}, normalizing
flows~\citep{shi2019graphaf,luo2021graphdf}, and Reinforcement Learning (RL)~\citep{yougr2019}. By
breaking the problem into smaller parts, these methods are more apt at
capturing complex structural patterns and can easily incorporate constraints
during generation. However, a key drawback of them is that their training is
sensitive to node ordering. Most existing works pre-define a fixed node
ordering by ad-hoc such as breadth-first search (BFS)
ordering~\citep{you2018graphrnn, shi2019graphaf}, which can be suboptimal for generation. OM \citep{chen2021order} also learns the node ordering for autoregressive graph generation, but our method differs from OM in two aspects.
(1) The motivations are different. Our method is motivated by autoregressive diffusion  and treats the graph sequences in the  diffusion process as the latent variable; OM treats the node ordering as the latent variable and infers its posterior distribution in a way similar to Variational autoencoder (VAE). (2) Our training objective is much simpler than OM. First, we do not need to compute the complicated graph automorphism, which requires some approximation algorithms to compute. Second, our training objective does not involve the entropy of the node ordering distribution. Existing works \citep{lucas2019don,lucas2019understanding} have shown that the VAE objective can cause posterior collapse.
\vspace{-1em}
\paragraph{Diffusion and Score-Based Generation.} Diffusion models have
emerged as a new family of powerful deep generative models. Denoising
diffusion probabilistic modeling (DDPM)~\citep{sohl-dicksteinde2015,ho2020denoising} perturbs the data
distribution into a Gaussian distribution through an forward Markov
noising process, and then learns to recover data distribution via the reverse
transition of the Markov chain. Closely related to DDPM is score-based
generation~\citep{songge2020}, which perturbs data with gradually increasing
noise, and then learns to reverse the perturbation via score matching.
\citet{songsc2021} generalize diffusion models to continuous-time diffusion
using forward and backward SDEs.
Existing diffusion-based graph generation models are all one-shot. \citet{niu2020permutation} model the adjacency matrices using score matching at different noise scales, and uses annealed Langevin dynamics  for generation; \citet{jo2022score} propose a continuous-time graph diffusion model that jointly models adjacency matrices and node features through stochastic differential equations (SDEs). Recently, \citet{vignac2022digress} introduce a discrete graph diffusion process by defining a Markov transition matrix for different node and edge types while \citet{haefeli2022diffusion} propose a similar model but only for the non-attributed graph. 
Different from these works, our model is the first \emph{autoregressive} diffusion model for graph generation, which defines diffusion directly in the discrete graph space. As we point out in Section~\ref{sec:intro}, our framework represents a generalized form of diffusion models, encompassing a broader class of models, including ARDM \cite{hoogeboom2022autoregressive} and diffusion Schrödinger bridge \cite{de2021diffusion}.

\section{Background}

\paragraph{Diffusion Model and Absorbing Diffusion}

Given a training instance $\vx_0 \in \sR^{D}$ sampled from the underlying distribution $p_{\rm data}(\vx_0)$, a diffusion model defines a forward Markov transition kernel $q(\vx_{t}|\vx_{t-1})$ to gradually corrupt training data until the data distribution is transformed into a simple noisy distribution.
The model then learns to reverse this process by learning a denoising transition kernel parameterized by a neural network $p_{\theta}(\vx_{t-1}|\vx_t)$.

Most existing works on diffusion models use Gaussian diffusion for
continuous-state data. To apply Gaussian diffusion on discrete data, one can
use the dequantization method by adding small noise to the data. However,
dequantization distorts the original discrete distribution, which can cause
difficulty in training diffusion-based models. For example, dequantization on
graph adjacency matrices can destroy graph connectivity information and hurt
message passing. \cite{austin2021structured, hoogeboom2021argmax} introduce
several discrete state space diffusion models using different Markov
transition matrices. Among them, absorbing diffusion is a promising one due to its simplicity and strong empirical performance.

% since the adding noise to the adjacency matrix can hurt the structural
% information of a graph.

\begin{definition}[Absorbing Discrete Diffusion] An absorbing diffusion is a Markov
  destruction process defined in the discrete state space. At transition time
  step $t$, each element $\vx_t^{(i)}$ in dimension $i$ is independently
  decayed into an absorbing state with probabilities $\alpha(t)$.
\end{definition}

The absorbing state can be a [MASK] token for texts or gray pixel for images
\citep{austin2021structured}. The diffusion process will converge to a
  stationary distribution that has all the mass on the absorbing state. The
  reverse of the absorbing diffusion is learned with categorical distribution
  to generate the original data. Typically, the decaying probabilities
  $\alpha(t)$ need to be small and the diffusion steps $T$ need to be large to
  attain good performance.

\vspace{-0.6em}
\paragraph{Autoregressive Diffusion Model}

Although absorbing diffusion directly operates on the discrete data, it still
needs a large number of diffusion steps and thus its generation process can be
slow. Autoregressive diffusion model (ARDM)
\citep{hoogeboom2022autoregressive} describes a fixed absorbing diffusion
process where the generative steps is equal to the dimensionality of the data,
\eg, the number of pixels in an image or the number of tokens in a piece of
text.

\begin{definition}[Autoregressive Diffusion]
  An autoregressive diffusion is a stochastic absorbing process where exactly one dimension decays to the absorbing state at a time. The  diffusion is repeated until all dimensions are absorbed.
\end{definition}

An equivalent way to describe this process is to first sample a permutation
$\sigma\in S_D$, where $S_D$ represents the set of all permutations of the
dimension indices $1,\cdots, D$. Then each dimension of the data decays in
that order towards the absorbing state. The corresponding generative process
then models the variables in the \emph{exact opposite order} of the
permutation. ARDM amounts to absorbing diffusion with a continuous time limit, as
detailed in Appendix \ref{app:arm}.

While ARDM offers an efficient and general diffusion framework for discrete
data, two key questions remain to be addressed for applying ARDM for graphs:
(1) How do we define absorbing states for inter-dependent nodes and edges in
graphs without losing the efficiency of ARDM? (2) While ARDM imposes a uniform
ordering for arriving at an order-agnostic variational lower bound (VLB) of
likelihood, a random ordering fails to capture graph topology. How do we
obtain a data-dependent ordering that leverages graph structural regularities
during generation? In the next section, we address these two challenges in our
proposed model.

\section{Method}

\begin{figure*}[t]
\centering
  \includegraphics[width=0.9\linewidth]{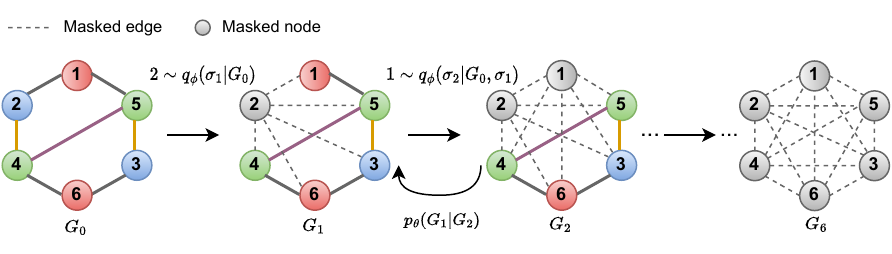}
  \vspace{-1.5em}
  \caption{The autoregressive graph diffusion process. In the forward pass,
    the nodes are autoregressively decayed into the absorbing states, dictated by an ordering generated by the diffusion ordering network $q_{\phi}(\sigma|G_0)$. In the reverse pass, the generator network $p_{\theta}(G_t|G_{t+1})$ reconstructs the graph structure using the reverse node ordering. Note that we do not need to consider the graph automorphism as \citep{chen2021order}, since the diffusion process assigns a unique ID to each node in $G_0$ to obtain the decay ordering. Therefore, there is a one-to-one mapping between $G_{0:n}$ and $\sigma_{1:n}$.
    For example, $v_1$  and $v_6$ have the same topology, but the denoising network will recover the exact node $v_1$ at $t=2$ since $\sigma_2=1$. We provide more illustrations in Appendix.~\ref{sec:appendix:comparison} }
  \label{fig:overall}
  \vspace{-1em}
\end{figure*}

\begin{figure*}
\centering
  \includegraphics[width=0.9\linewidth]{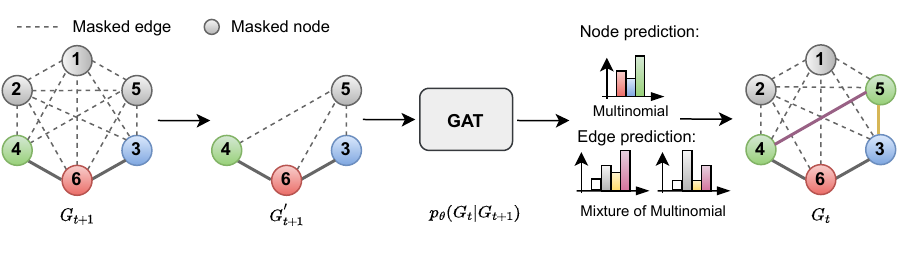}
  \vspace{-2em}
  \caption{
    The generation procedure at step $t$ with the denoising network $p_{\theta}(G_t|G_{t+1})$. The denoising network predicts the node type of node $v_{\sigma_t}$ and its edges with all previously denoised nodes. 
  }
  \label{fig:denosing}
  \vspace{-1em}
\end{figure*}

A graph is represented by the tuple $G=(V,E)$ with node set $V=\{v_1,\cdots,v_n\}$ and edge set $E=\{e_{v_i,v_j}|v_i,v_j\in V) \}$. We denote by $n=|V|$ and $m=|E|$ the number of nodes and edges in $G$ respectively. Each node and edge have their corresponding categorical labels, \eg, $e_{v_i, v_j}=k$ represents that the edge between node $v_i$ and $v_j$ is in type $k$. We treat the absence of edges between two nodes as a particular edge type.
Our goal is to learn a graph generative model  from a set of training graphs.

\subsection{Autoregressive Graph Diffusion Process}

Due to the dependency between nodes and edges, it is nontrivial to apply
absorbing diffusion \citep{austin2021structured} in the discrete graph space.
We first define  absorbing node state on graphs as follows:

\begin{definition}[Absorbing Node State]
   When a node $v_i$ enters the absorbing state, (1) it will be masked and (2)
  it will be connected to all the other nodes in $G$ by masked edges.
\end{definition}

% In practice, we can use an
% additional [MASK] class to represent the masked edges so that the model can
% distinguish them from the original edges.

% \bap{Maybe we can be more specific and say Why node ordering matters? give an example  with two different orderings?}

Instead of only masking the original edges, we connect the masked node $v_i$
to all the other nodes with masked edges as we cannot know $v_i$'s
original neighbors in the absorbing state. With the absorbing node state defined, we then need a node decay ordering for the forward absorbing pass.
A na\"ive strategy is to use a random ordering sampled from a uniform distribution as in \citep{hoogeboom2022autoregressive}.
In the reverse generation process, the variables will be generated in the exact reverse order, which also follows a uniform distribution.
However, such a strategy is problematic for graphs.
First, different graph datasets have different structural regularities, and it is key to leverage such regularities to ease generative learning.
For example, community-structured graphs typically consist of  dense subgraphs that are loosely overlapping.
For such graphs, it is an easier learning task to generate one community first and then add the others, but a random node ordering cannot leverage such local structural regularity, which makes generation more difficult.
Second,  to compute the likelihood, we need to marginalize over all possible node orderings  due to node permutation invariance.
It will be more sample efficient if we can use an optimized proposal ordering distribution and use importance sampling to compute the data likelihood.

% While in our method, we can use the diffusion ordering network as an optimized proposal distribution for importance sampling and thus it is more sample efficient.

To address this issue, we propose to use a
diffusion ordering network $q_{\phi}(\sigma|G_0)$ such that, at each diffusion step $t$,
we sample from this network to select a node $v_{\sigma(t)}$ to be absorbed and obtain the corresponding masked graph $G_t$ (Figure~\ref{fig:overall}). This leads to
the following definition of our graph autoregressive diffusion process:

\begin{definition}[Autoregressive Graph Diffusion Process]
  In autoregressive graph diffusion, the node decay ordering
  $\sigma$
  is sampled from a diffusion ordering network $q_{\phi}(\sigma|G_0)$. Then,
  exactly one node decays to the absorbing state at a time according to the sampled diffusion ordering. The process proceeds until all the nodes are absorbed.
\end{definition}

The diffusion ordering network follows a recurrent structure $q_{\phi}(\sigma|G_0)=\prod_{t}q_{\phi}(\sigma_t|G_0,\sigma_{(<t)})$.
At each step $t$, the distribution of the $t$-th node $\sigma_t$ is conditioned on the original graph $G_0$ and the generated node ordering up to $t-1$, \ie $\sigma_{({<t})}$.
We use a graph neural network (GNN) to encode the structural information in the graph.
To capture the partial ordering, we add positional encodings into  node features \citep{vaswani2017attention} as in \citep{chen2021order}.
We denote the updated node embedding of node $v_i$ after passing the GNN as $\vh_i^d$, and parameterize $q_{\phi}(\sigma_t|G_0,\sigma_{(<t)})$ as a categorical distribution:
\begin{equation}
  q_{\phi}(\sigma_t|G_0,\sigma_{(<t)})=\frac{\text{exp}(\vh_i^d)}{\sum_{i^{'}\notin \sigma_{(<t)}}\text{exp}(\vh^d_{i^{'}})}.
\end{equation}
With $q_{\phi}(\sigma|G_0)$, \ours can learn to optimize node ordering for
diffusion. However, this also requires us to infer the reverse generation
ordering in the backward pass. Inferring such a reverse generation ordering
is difficult since we do not have access to the original graph $G_0$ in
intermediate backward steps. In Section~\ref{sec:training}, we show that it
is possible to circumvent inferring this generation ordering by leveraging
the permutation invariance of graph generation.

\subsection{The Reverse Generative Process}

In the generative process, a denoising network $p_{\theta}(G_t|G_{t+1})$ will denoise the masked graph in
the reverse order of the diffusion process.  We
design
$p_{\theta}(G_t|G_{t+1})$
as a graph attention network (GAT) \citep{velickovic2018graph,liao2019efficient} parameterized by
$\theta$,
so that the model can distinguish the masked and
unmasked edges. For clarity, we use the Vanilla GAT to illustrate the computing process. 
However, one can adopt any advanced graph neural network with attentive message passing.

% along with its node features. Since we
% focus on non-attributed graphs, we use the node degree and its one-hot
% encoding vector as the feature.

At time $t$, the input to the denoising network $p_{\theta}(G_t|G_{t+1})$ is
the previous masked graph $G_{t+1}$. A direct way is to use $G_{t+1}$ which
contains all the masked nodes with their corresponding masked edges. However,
during the initial generation steps, the graph is nearly fully connected with
masked edges. This has two issues: (1) the message passing procedure will be
dominated by the masked edges which makes the messages uninformative. (2)
Storing the dense adjacency matrix is memory expensive, which makes the model
unscalable to large graphs. Therefore, during each generation step, we only
keep the masked node to be denoised with its associated masked edges, while
ignoring  the other masked nodes. We refer the modified masked graph as
$G^{'}_t$, as shown in Figure~\ref{fig:denosing}.

The denoising network first uses an embedding layer to encode each node $v_i$ into a continuous embedding space, \ie ~$\vh_i=\text{Embedding}(v_i)$.
At $l$-th message passing, we update the embedding of node $v_i$ by aggregating the attentive messages from its neighbor nodes:
%\begin{align}
%  \alpha_{i,j}&=\frac{\text{exp}(\text{LeakyReLU}(\va^T[\mW\vh_i||\mW\vh_j])}{\sum_{k\in\mathcal{N}_i}\text{exp}(\text{LeakyReLU}(\va^T)[\mW\vh_i||\mW\vh_j])}, \nonumber\\ \vh_i&=\text{ReLU}\left(\sum_{j\in\mathcal{N}_i}\alpha_{i,j}\mW \vh_j\right),
%  \label{eq:GAT}
%\end{align}
 $ \alpha_{i,j}=\frac{\text{exp}(\text{LeakyReLU}(\va^T[\mW\vh_i||\mW\vh_j])}{\sum_{k\in\mathcal{N}_i}\text{exp}(\text{LeakyReLU}(\va^T)[\mW\vh_i||\mW\vh_j])}, \vh_i=\text{ReLU}\left(\sum_{j\in\mathcal{N}_i}\alpha_{i,j}\mW \vh_j\right),$
where $\mW$ is the weight matrix, $\va$ is the attention vector.  The attention mechanism enables the model to distinguish if the message comes from a masked edge. After $L$ rounds of message passing, we obtain the final embedding $\vh_i^{L}$ for each node,  then we predict the node type of the new node $v_{\sigma_t}$ and the edge types between $v_{\sigma_t}$ and all previously denoised nodes $\{v_{\sigma(>t)}\}$. The node type prediction follows a multinomial distribution. For edge prediction,
one choice is to sequentially predict these edges as in \citep{you2018graphrnn,shi2019graphaf}. However, this sequential generation process is inefficient and takes $\mathcal{O}(n^2)$. Instead, we predict the connections of the new node to all previous nodes at once using a mixture of multinomial distribution.
The mixture distribution can capture the dependencies among edges to be generated and  meanwhile  reduce the autoregressive generation steps to $\mathcal{O}(n)$.

%Note that we do not need to compute the graph automorphism as \citep{chen2021order}. This is because, in our framework, the node ordering $\sigma_{1:n}$ is sampled in the forward diffusion process which essentially assigns a unique ID to each node of $G_0$. In the generation procedure, our starting point is the masked graph $G_n$ with $n$  nodes; our generator network sequentially denoises the node $v_{\sigma_t}$ in $G_n$ in the reverse order of the diffusion ordering during training. In contrast, the previous work can only sequentially grow the permuted adjacency matrix but such permuted matrix loses the exact node ordering information.
\subsection{Training Objective}
\label{sec:training}
We use approximate maximum likelihood as the training objective for \ours. We first derive the variational lower bound (VLB) of likelihood as:
\begin{align}
 \log{p_{\theta}(G_{0})} &=  \log{\left(\int p(G_{0:n})\frac{q(G_{1:n}|G_0)}{q(G_{1:n}|G_0)}dG_{1:n}\right)} \nonumber\\
                          &\geq \mathbb{E}_{q(\sigma_{1:n}|G_0)}
                            \sum_{t}\log{p_{\theta}(G_t|G_{t+1})}\nonumber\\
                        &\quad\,-\text{KL}(q_{\phi}(\sigma_{1:n}|G_0)|p_{\theta'}(\sigma_{1:n}|G_n)),
                            \label{eq:VLB}
\end{align}
where $G_{0:n}$ denotes all values of $G_t$ for $t=0,\cdots,n$ and $p_{\theta'}(\sigma_{1:n}|G_n)$ is the distribution of the generation ordering. A detailed derivation of Eq.~\ref{eq:VLB} is given in Appendix \ref{sec:appendix:derivation}

As we can see from Eq.~\ref{eq:VLB},  the diffusion process introduces a separate reverse generation ordering network $p_{\theta'}(\sigma_{1:n}|G_n)$. Learning $p_{\theta'}(\sigma_{1:n}|G_n)$ is nontrivial as we do not have the original graph $G_0$ in the intermediate generation process.  However, we show that we can avoid such difficulty and simply
ignore the KL-divergence term.  
While the generation ordering network is required for non-graph data such as text to determine which token to unmask at test time, it is not needed for graph generation due to node permutation invariance. The first term will encourage the denoising network $p_{\theta}(G_t|G_{t+1})$
to predict the node and edge types in the exact reverse ordering of the diffusion process, thus the denoising network itself can be a proxy of the generation ordering.
Due to permutation invariance, we can simply replace any masked node and its masked edges with the predicted node and edge types at each time step.    Therefore, we can ignore the second term and finally arrive at a simple training objective:
\begin{align}
  L_{\rm train}& =\mathbb{E}_{\sigma_{1:n} \sim q_{\phi}(\sigma_{1:n}|G_0)} \sum_{t} p_{\theta}(G_t|G_{t+1})  \\ \nonumber
  &= n\mathbb{E}_{\sigma_{1:n} \sim q_{\phi}(\sigma_{1:n}|G_0)} \mathbb{E}_{t\sim \mathcal{U}_n} p_{\theta}(O_{v_{\sigma_t}}^{\sigma_{(>t)}}|G_{t+1}),
  \label{eq:loss}
\end{align}
where the last equivalence comes from treating $t$ as the random variable with a uniform distribution $\mathcal{U}_n$ over $1$ to $n$. $O_{v_{\sigma_t}}^{\sigma_{(>t)}}$ represents the node type of $v_{\sigma_t}$ and its edges with all previously denoised nodes, \ie~$\{v_{\sigma_t}, \{e_{v_{\sigma_t},v_j}\}_{j=\sigma_{t+1}}^{\sigma_n}\}$. 

Compared with the random diffusion ordering, our design has two benefits: (1) We can automatically learn a data-dependent node generation ordering which leverages the graph structural information. (2) We can consider the diffusion ordering network as an optimized proposal distribution of importance sampling for computing the data likelihood, which is more sample-efficient than a uniform proposal distribution.

\vspace{-1em}
\paragraph{Soft Label Training} The architecture of ARDM \citep{hoogeboom2022autoregressive} predicts all masked dimensions simultaneously, which 
enables training the univariate conditionals $p(\vx_k|\vx_{\sigma(>t)})$ for all $k \in \sigma_{(\leq t)}$  in parallel. In \ours, due to the node permutation invariant property of graph,
this parallel training can be simplified as training with a soft label which is weighted by the probability given by the diffusion ordering network:
\begin{align}
L_{\rm train}= n\mathbb{E}_{q_{\phi}(\sigma_{1:n}|G_0)}\mathbb{E}_{t\sim \mathcal{U}_n}\sum_{k\in \sigma_{(\leq t)}}w_k p_{\theta}(O_{v_{k}}^{\sigma_{(>t)}}|G_{t+1}),\nonumber
  %\label{eq:parallel}
\end{align}
where $w_k = q_{\phi}(\sigma_t=k|G_0, \sigma_{(<t)})$. 
In practice, the probability mass $q_{\phi}(\sigma_t|G_0, \sigma_{(<t)})$ might be concentrated around a small set of remaining nodes. Therefore, it is generally sufficient to consider only those node labels associated with the highest probabilities.

\setlength{\tabcolsep}{5pt}

\subsection{Parameter Optimization}

Learning  the parameters of \ours is challenging, because we need to evaluate
the expectation of the likelihood over the diffusion ordering network. We use
a reinforcement learning (RL) procedure by sampling multiple diffusion
trajectories, thereby enabling training both the diffusion ordering network
$q_{\phi}(\sigma|G_0)$ and the denoising network $p_{\theta}(G_t|G_{t+1})$ using
gradient descent.

Specifically, at each training iteration, we explore the diffusion ordering
network by creating $M$ diffusion trajectories for each training graph $G^{(i)}_0$. Each trajectory is a sequence of graphs $\{G_t^{i,m}\}_{1\leq t\leq n}$ where the node decay ordering $\sigma^{i,m}$ is sampled from $q_{\phi}(\sigma|G_0^{i,m})$. For each trajectory, we sample $T$ time steps.
The denoising network
$p_{\theta}(G_t|G_{t+1})$ is then trained to minimize the negative VLB using stochastic gradient descent (SGD):
\begin{align}
\triangle \theta \leftarrow
  \frac{\eta_1}{M}\nabla \sum_{i\in\mathcal{B}_{\text{train}}}\sum_{ m,t}\sum_{k\in\sigma_{(\leq t)}}\frac{n_iw^{i,m}_{k}}{T}\log p_{\theta}(O_{v_{k}}^{\sigma_{(>t)}}|G^{i,m}_{t+1}), \nonumber
\end{align}
where $\mathcal{B}_{\rm train}$ is the a minibatch sampled from the training data and $w_k^{i,m}=q_{\phi}(\sigma_t^{i,m}=k|G_0^i, \sigma_{(<t)}^{i,m})$.

To evaluate the current
diffusion ordering network, we create $M$ trajectories for each validation graph and
compute the negative VLB of the
denoising network to obtain the corresponding rewards $R^{i,m}=-\sum_{t}\sum_{k\in \sigma_{(\leq t)}}\frac{n_i}{T}w_{k}^{i,m}\text{log} p_{\theta}(O_{v_{k}}^{\sigma_{(>t)}}|G^{i,m}_{t+1})$. Then, the diffusion ordering network can be updated with common RL optimization methods, \eg, the REINFORCE algorithm \citep{williams1992simple}:
\begin{equation}
  \triangle\phi \leftarrow \frac{\eta_2}{M} \sum_{i\in \mathcal{B}_{\text{val}}}\sum_{m} R^{i,m} \nabla \log q_{\phi}(\sigma|G_0^{i,m}).
  \label{eq:reinforce}
\end{equation}
The detailed training procedure is summarized in Algorithm~\ref{alg:overall} in Appendix.~\ref{sec:appendix:alg}.

\begin{table*}[t]
  \centering
  \scriptsize
  \scalebox{1.2}{
      \begin{tabular}{@{}lcccccccccccccc@{}}
        \toprule
        \multirow{2}{*}{\textbf{Model}} & \multicolumn{4}{c}{\textbf{Community-small}} & \phantom{c} & \multicolumn{4}{c}{\textbf{Caveman}} & \phantom{c} & \multicolumn{4}{c}{\textbf{Cora}} \\ \cmidrule{2-5} \cmidrule{7-10} \cmidrule{12-15}
          & \textbf{Deg.} & \textbf{Clus.} & \textbf{Orbit} & \textbf{Time/s} & \phantom{c} & \textbf{Deg.}   & \textbf{Clus.} & \textbf{Orbit} & \textbf{Time/s} & \phantom{c} & \textbf{Deg.}   & \textbf{Clus.} & \textbf{Orbit} & \textbf{Time/s} \\ \midrule
        DeepGMG & 0.220 & 0.950 & 0.400 & 496.6 & \phantom{c} & 1.752 & 1.642 & 0.2122 & 530.2 & \phantom{c} & - & - & - & - \\
        GraphRNN & 0.080 & 0.120 & 0.040 & 16.4 & \phantom{c} & 0.371 & 1.035 & 0.033 & 27.0 & \phantom{c} & 1.689 & 0.608 & 0.308 & 33.3 \\
        GraphAF& 0.180 & 0.200 & 0.020 & 19.3 & \phantom{c} & 0.269 & 0.587 & 0.422 & 20.5 & \phantom{c} & 0.176 & \underline{0.080} & \underline{0.094} & 108.5 \\
        GraphDF & 0.060 & 0.120 & 0.030 & 10.3 & \phantom{c} & 0.077  & 0.373 & 0.051  & 18.5 & \phantom{c} & 0.454  & \textbf{0.074} & 0.256  & 129.7 \\
        GraphVAE & 0.350 & 0.980  & 0.540 & 0.2 & \phantom{c} & 1.402  & 1.086    & 1.391  & 0.2 & \phantom{c} & 1.521  & 1.740    & 0.788  & 0.3 \\
        GRAN    & 0.060 & 0.110  & 0.050 & 1.8 & \phantom{c} & 0.043 & 0.130  & 0.018 & 2.5 & \phantom{c} & \underline{0.125}    & 0.272  & 0.127 & 5.1     \\
        OM        & 0.047 & 0.130  & 0.008      & 2.0   & \phantom{c} & 0.032  & 0.076  & 0.027   & 3.0 & \phantom{c} & 0.249  & 0.201    & 0.145  & 5.6 \\
        EDP-GNN   & 0.053 & 0.144  & 0.026 & 2.9$e^3$ & \phantom{c} & 0.032    & 0.168  & 0.030 & 1.8$e^3$ & \phantom{c} & \textbf{0.093} & 0.269  & 0.062 & 4.6$e^3$ \\
                      SPECTRE  & 0.048 & 0.049   & 0.016  & 0.4   & \phantom{c}     & \textbf{0.013}  & 0.084   &0.028   & 0.5 & \phantom{c} & 0.021  & \underline{0.080} & \textbf{0.007}  & 0.14 \\
        GDSS   & \underline{0.045}    & 0.086   & \underline{0.007}    & 4.3$e^2$ & \phantom{c} & \underline{0.019} & 0.048 & \textbf{0.006} & 5.1$e^2$       & \phantom{c} & 0.160 & 0.376  & 0.187 & 4.2$e^2$ \\
                DiGress  & 0.047 & \textbf{0.041}   & 0.026  & 5.7   & \phantom{c} & \underline{0.019}  & \underline{0.040}   & \textbf{0.003}  & 10.4 & \phantom{c} & 0.044 & 0.042 & 0.223  & 79.7 \\
        \rowcolor[HTML]{EFEFEF}\ours   & \textbf{0.034} & \underline{0.082}      & \textbf{0.004} & 1.9 & \phantom{c} & 0.039 & \textbf{0.028}   & 0.018    & 2.7 & \phantom{c} & 0.273 & 0.138  & 0.105    & 4.5  \\ \midrule
        \multirow{2}{*}{\textbf{Model}} & \multicolumn{4}{c}{\textbf{Breast}} & \phantom{c} & \multicolumn{4}{c}{\textbf{Enzymes}} & \phantom{c} & \multicolumn{4}{c}{\textbf{Ego-small}} \\ \cmidrule{2-5} \cmidrule{7-10} \cmidrule{12-15}
        & \textbf{Deg.}   & \textbf{Clus.} & \textbf{Orbit} & \textbf{Time/s} & \phantom{c} & \textbf{Deg.}   & \textbf{Clus.} & \textbf{Orbit} & \textbf{Time/s} & \phantom{c} & \textbf{Deg.} & \textbf{Clus.} & \textbf{Orbit} & \textbf{Time/s} \\ \midrule
        DeepGMG & -  & -    & -  & - & \phantom{c} & -  & -    & -  & - & \phantom{c} & 0.040  & 0.100    & 0.020  & 477 \\
        GraphRNN & 0.103 & 0.138  & 0.005 & 31.0  & \phantom{c} & \underline{0.017}  & 0.062    & 0.046  & 19.2 & \phantom{c} & 0.090  & 0.220    & 0.003  & 18.7\\
        GraphAF & 0.111  & 0.407    & 0.003  & 53.1 & \phantom{c} & 1.669  & 1.283    & 0.266  & 28.6  & \phantom{c} & 0.030  & 0.110   & \textbf{0.001}  & 6.9 \\
        GraphDF & 0.283  & 0.078    & 0.035  & 62.4 & \phantom{c} & 1.503  & 1.061    & 0.202  & 39.8 & \phantom{c} & 0.040  & 0.130    & 0.010  & 10.2 \\
        GraphVAE  & 1.591 & 1.993  & 1.050 & 0.2  & \phantom{c} & 1.369  & 0.629 & 0.191  & 0.7 & \phantom{c} & 0.130 & 0.170  & 0.050   & 0.3 \\
        GRAN & 0.073 & 0.413  & 0.010 & 2.1  & \phantom{c} & 0.054  & 0.078  & 0.017 & 3.2 & \phantom{c}  & 0.030  & 0.029    & 0.014  & 1.2 \\
        OM & \underline{0.042}    & 0.140  & 0.005 & 2.8  & \phantom{c} & 0.051 & 0.083 & 0.024  & 2.4 & \phantom{c} & 0.024  & 0.035    & 0.018  & 1.4\\
        EDP-GNN  & 0.131 & \underline{0.038}      & 0.019 & 3.6$e^3$ & \phantom{c} & 0.023  & 0.268    & 0.082  & 2.2$e^3$      & \phantom{c} & 0.052 & 0.093 & 0.007   & 3.9$e^3$     \\
                SPECTRE  & 0.312 & 0.837   & 0.087  & 0.14  & \phantom{c}     &  0.136 &  0.195 &   0.125 & 0.9& \phantom{c} &   0.078 & 0.078  & 0.007 & 0.4\\
        GDSS  & 0.113 & \textbf{0.020}   & \underline{0.003}    & 8.6$e^2$   & \phantom{c}     & 0.026  & \underline{0.061}    & \underline{0.009}  & 4.4$e^2$  & \phantom{c} & \underline{0.021} & \underline{0.024} & 0.007  & 2.6$e^2$ \\
        DiGress  & 0.152 & 0.024   & 0.008  & 91.5   & \phantom{c}     & \textbf{0.004}  & 0.083   & \textbf{0.002}  & 89.48 & \phantom{c} & \textbf{0.015}  & 0.029 & \underline{0.005}  & 8.8\\
        
        \rowcolor[HTML]{EFEFEF}\ours & \textbf{0.036} & 0.041  & \textbf{0.002} & 2.3  & \phantom{c} & 0.029 & \textbf{0.054} & 0.015 & 3.5 & \phantom{c} & \underline{0.019}  & \textbf{0.017}   & 0.010  & 1.2 \\\bottomrule      \end{tabular}}
  \caption{Generation results on the six generic graph datasets for all the methods. Best results are bold and the second best values are underlined (smaller the better). ``-'' denotes out-of-resources that take more than 10 days to run. 
  }
  \label{table:main}
\end{table*}

\vspace{-0.5em}
\section{Experiments}

\subsection{Generic Graph Generation}

\textbf{Experimental Setup}.
We evaluate the performance of \ours on six diverse graph generation benchmarks from different domains:
(1) Community-small \citep{you2018graphrnn},
(2) Caveman \citep{You},
(3) Cora \citep{sen2008collective},
(4) Breast \citep{gonzalez2011high},
(5) Enzymes \citep{schomburg2004brenda} and
(6) Ego-small \citep{sen2008collective}. For each dataset, we use $80\%$ of the graphs as training set and the rest $20\%$ as test sets. Following \citep{liao2019efficient}, we randomly select $20\%$ from the training data as the validation set. We generate
the same amount of samples as the test set for each dataset.
More details can be seen in Appendix.\ref{sec:appendix:dataset}.
Following previous work~\citep{you2018graphrnn}, we measure generation quality using the maximum
mean discrepancy (MMD) as a distribution distance between the generated
graphs and the test graphs. Specifically, we compute the MMD of degree
distribution, clustering coefficient, and orbit occurrence numbers of 4 nodes between
the generated set and the test set. 
We also report the
generation time of different methods.

\textbf{Baselines and Implementation Details}.  
We compare \ours with the following baselines: DeepGMG~\citep{li2018learning} and GraphRNN~\citep{you2018graphrnn} are RNN-based autoregressive graph generation models. GraphAF~\citep{shi2019graphaf} and GraphDF~\citep{luo2021graphdf} are flow-based autoregressive models. GRAN~\citep{liao2019efficient} is an autoregressive model conditioned on blocks. OM~\citep{chen2021order} is an autoregressive model that infers node ordering using variational inference. GraphVAE~\citep{simonovsky2018graphvae} 
is an one-shot model based on VAE. SPECTRE \citep{martinkus2022spectre} is a one-shot model based on GAN.
DiGress \citep{vignac2022digress},
EDP-GNN~\citep{niu2020permutation} and 
GDSS~\citep{jo2022score} are 
diffusion-based one-shot methods. Implementation details and parameter settings are provided in in Appendix \ref{sec:appendix:implementation}.

\vspace{-1em}
\paragraph{Experimental Results}

Table~\ref{table:main} shows the generation performance on the six benchmarks for all the methods. (1) As we can see, our method can outperform or achieve competitive performance compared with the baselines. In terms of efficiency, our model is on par with the most efficient autoregressive baseline  GRAN, and 10-100X faster than other baselines except for the one-shot baselines GraphVAE and SPECTRE. However, GraphVAE's generation quality is much worse  than ours while SPECRE is a GAN based model and cannot provide likelihood computation.
(2) GDSS and DiGress are the strongest baselines. However, their generation speeds are extremely slow as they need a long diffusion process to arrive at the stationary distribution. Our \ours is up to 100X times faster than GDSS and 30X times faster than DiGress.
The other graph diffusion model EDP-GNN is even slower as its generation process uses annealed Langevin Dynamics with many different noise levels. (3) \ours also outperforms existing autoregressive graph generative models by large margins. This is because they adopt a fixed node ordering when training the generative model while \ours automatically learns a data-dependent generation ordering for the graph. 
 Though OM also learns a node ordering distribution,  \ours consistently outperforms it. This is because OM uses the VAE objective which may suffer from posterior collapse, while \ours has a much simpler training objective based on autoregressive diffusion.

\begin{table*}[!t]
\centering
\footnotesize
\setlength{\tabcolsep}{0.3em}
\begin{tabular}{@{}lcccccc|cccccc@{}}
\toprule
         & \multicolumn{6}{c}{\textbf{QM9}}                                    & \multicolumn{6}{|c}{\textbf{ZINC250k}}                            \\ \cmidrule(l){2-13} 
\textbf{Model}   & Validity$\uparrow$ & NSPDK$\downarrow$  & FCD$\downarrow$    & Unique$\uparrow$ & Novelty$\uparrow$ & Time$\downarrow$   & Validity$\uparrow$ & NSPDK$\downarrow$ & FCD$\downarrow$    & Unique$\uparrow$ & Novelty$\uparrow$ & Time$\downarrow$ \\ \midrule
GraphAF  & 74.43    & 0.020  & 5.27  & 88.64      & 86.59   & 3.01$e^3$   & 68.47    & 0.044 & 16.02 & 98.64      & 100     &  $7.2e^3$    \\
GraphDF  & 93.88    & 0.064  & 10.93 & 98.58      & 98.54   & 5.8$e^4$  & 90.61    & 0.177 & 33.55 & 99.63      & 100     & $6.95e^4$     \\
MoFlow   & 91.36    & 0.017  & 4.47  & 98.65      & 94.72   & 5.43   & 63.11    & 0.046 & 20.93 & 99.99      & 100     &  36.1     \\
EDP-GNN  & 47.52    & 0.005  & 2.68  & 99.25      & 86.58   & 52.$e^2$  & 82.97    & \underline{0.049} & 16.74 & 99.79      & 100     &   $1.21e^4$   \\
GraphEBM & 8.22     & 0.030  & 6.14  & 97.90      & 97.01   & 44.2 & 5.29     & 0.212 & 35.47 & 98.79      & 100     &  87.2    \\
SPECTRE  & 87.3     & 0.163  & 47.96 & 35.7       & 97.28   & 3.3    &    90.2      &  0.109    & 18.44       & 67.05            &  100       & 103.1     \\
GDSS     & \underline{95.72}    & 0.003  & 2.9    & 98.46      & 86.27   & 1.3$e^2$  & \textbf{97.01}    & \textbf{0.019} & \textbf{14.66} & 99.64      & 100     &  $2.58e^3$    \\
DiGress  & \textbf{99.0}     & \textbf{0.0005} & \textbf{0.36}  & 96.66      & 33.4    & 1.02$e^2$ &     \underline{91.02}     &   0.082    & 23.06       &  81.23          &      100   &   1.52$e^3$   \\
\rowcolor[HTML]{EFEFEF} \ours& 90.25    & \underline{0.002}  & \underline{1.22}   & 95.62      & 70.39   & 15.2   &     88.23     &  0.055     &  \underline{16.26}      & 99.46           & 100        &    132.8  \\ \bottomrule
\end{tabular}
\caption{Generation results on the QM9 and ZINC250k dataset. Best results are bold and the second best values are underlined.}
%\vspace{-1em}
\label{table:molecule}
\end{table*}

\vspace{-0.5em}
\subsection{Molecule Generation}

\textbf{Experimental Setup} 
We use two molecular dataset, QM9 \citep{ramakrishnan2014quantum} and ZINC250k \citep{irwin2012zinc}.  Following previous works \citep{jo2022score}, we evaluate 10,000 generated molecules with the following metrics.
\textbf{Fréchet ChemNet Distance
(FCD)} \citep{preuer2018frechet} evaluates the distance between
the training and generated sets using the activations of the
penultimate layer of the ChemNet. \textbf{Neighborhood subgraph pairwise distance kernel (NSPDK) MMD } \cite{costa2010fast} is the MMD between the generated
molecules and test molecules which takes into account both
the node and edge features for evaluation. \textbf{Validity} is the fraction of valid molecules without valency correction. \textbf{Uniqueness} is the fraction of the valid molecules that are unique. \textbf{Novelty} is the fraction of the valid molecules
that are not included in the training set.
We provide the implementation details in Appendix.~\ref{sec:appendix:implementation:molecule}.

\textbf{Experimental Results}
Table.~\ref{table:molecule} shows the generation performance on the two molecule datasets for all the methods. \ours can achieve competitive performance with the strongest baseline while being much more efficient for generation. \ours is performing well, particularly on NSPDK and FCD which are two salient metrics measuring how close the generated molecules lie to the distribution in graph structure space and chemical space. On QM9, only DiGress outperforms \ours. However, its novelty score is only below 40\%. This indicates that DiGress may suffer from simply memorizing the training data. On ZINC250K, only GDSS outperforms \ours but 10X slower for generation. In terms of efficiency, only MoFlow and SPECTRE are faster than \ours but their generation performance is not on par with \ours. For the validity, though \ours is not the best one, this can be easily addressed using edge resampling during generation since \ours is an autoregressive model which is more flexible. The outstanding performance of \ours on molecule datasets verifies the effectiveness of our method for learning the underlying distribution of graphs with multiple node and edge types.

\subsection{Constrained Graph Generation}
 In this set of experiments, we study \ours's capability in constrained graph generation by comparing it with the strongest one-shot and autoregressive baselines. We use the Caveman dataset for
constrained graph generation, with the constraint that the maximum node degree
is no larger than 6. The detailed setup is in Appendix \ref{sec:appendix:constrained}.
Table.~\ref{table:constraints} shows the constrained generation performance on the Caveman dataset. 
We find that more than half of their generated samples are invalid. For autoregressive baselines, we apply the degree-checking procedure on the two strongest baselines, \ie GRAN and OM. DiGress, GDSS and EDP-GNN are all one-shot generative models and they are hard to incorporate such constraints during the generation procedure. Though it is possible to apply a post-hoc modification for DiGress, GDSS and EDP-GNN, such a strategy has not been explored in the existing literature for general graph generation.
As we can see, \ours can generate constrained samples that are closer to the data distribution. This is
useful for many real-world applications. For example, when designing the
contact networks of patients and healthcare workers in hospitals, a constraint
of degrees for healthcare workers can help  avoid superspreaders and
potential infectious disease
outbreaks~\citep{jang2019evaluating,adhikari2019fast}.

 \begin{table}[h]
  \centering
  \small
    \setlength{\tabcolsep}{1mm}{
  \begin{tabular}{@{}lllll@{}}
    \toprule
    Method    & Validity & Deg. & Clus. & Orbit \\ \midrule
    EDP-GNN &   $39\%$       &   -  &     - &     -  \\
    GDSS    &      $34\%$    & -    &    -  &  -     \\
    DiGress    &      $33\%$    & -    &    -  &  -     \\
    GRAN    &   $100\%$       & $0.208$    &   $0.231$   &    $0.158$   \\
    OM     &   $100\%$       & $0.190$    & $0.168$ & $\bm{0.132}$      \\
    \rowcolor[HTML]{EFEFEF}\ours    &   $100\%$       &  $\bm{0.176}$   & $\bm{0.157}$      & $ 0.144$       \\ \bottomrule
  \end{tabular}}
  \caption{Constrained graph generation results on the Caveman dataset.}
  \vspace{-1em}
  \label{table:constraints}
\end{table}

\begin{table}
\centering
\small
\begin{tabular}{@{}lllll@{}}
\toprule
    Method   & Deg.  & Clus. & Orbit & Counts\\\cmidrule(l){2-5} 
OA-ARDM & $0.085$ & $0.129$ & $0.032$ & $5.6$            \\
\rowcolor[HTML]{EFEFEF} \ours    & $\bm{0.034}$ & $\bm{0.082}$ & $\bm{0.004}$ & $\bm{2.6}$ \\ \bottomrule          
\end{tabular}
  \caption{Generation performance and node generation ordering of \ours and OA-ARDM on the Community-small dataset. Counts represents the average number of nodes that cross different clusters during the generation procedure. }
  \label{table:oa}
  \vspace{-1em}
\end{table}

\subsection{Ablation Study: Effect of Diffusion Ordering}

We further validate the superiority of the learned diffusion
ordering by comparing \ours with an ablation OA-ARDM, which uses random node ordering \citep{hoogeboom2022autoregressive} for graph generation. Table~\ref{table:oa}
shows the generation performance and generation ordering of \ours and OA-ARDM on the Community-small dataset. As shown, with the random generation ordering, the
generation performance drops significantly.  
To evaluate the node generation ordering, we use the spectral graph clustering method to partition the nodes into two clusters for each generated graph and then count the cross-cluster steps during the generation procedure. 
As we can see, the average number of nodes that cross different clusters of \ours is much smaller than OA-ARDM. This demonstrates that \ours tends first to generate one cluster and then adds another cluster while OA-ARDM just randomly generates the graph. Therefore, the generation ordering of \ours can better capture  graph topology regularity  than OA-ARDM. Figure~\ref{fig:case_study} (in Appendix \ref{sec:appendix:case}) visualizes the graph generation process. As we can see, \ours first generates one community and then moves to another; while OA-ARDM randomly generates the graph and fails to capture the underlying graph distribution. 

We remark that the learned node orderings can only be transferred when the two datasets share structural similarities, as our learned node ordering is data-dependent. For instance, transferability is possible if both datasets represent social communities with similar structural properties, e.g., several dense subgraphs that are loosely overlapping. However, if the datasets do not exhibit such similarities, the node ordering cannot be effectively transferred.

\vspace{-2mm}
\section{Limitations and Discussion}
\vspace{-2mm}

We have proposed a new autoregressive diffusion model for graph generation.
The proposed model \ours defines a node-absorbing diffusion process that
directly operates in the discrete graph spaces. We designed a diffusion
ordering network that learns a data-dependent ordering for this diffusion process,
coupled with a reverse denoising network that performs autoregressive graph
reconstruction. We derived a simple variational lower bound of the likelihood,
and showed that the two networks
can be jointly trained with reinforcement
learning. Our experiments have validated the generation performance and
efficiency of \ours. 

We discuss limitations and possible extensions of \ours: 
(1) \emph{Handling more complex constraints.} We have shown that the
autoregressive  procedure of \ours can handle constraints better
than one-shot generative models. However, practical graph generation
applications can involve complex constraints on global-level properties such
as number of articulation points, which is 
challenging for our model and existing graph generative models. (2) \emph{Memory and time efficiency.} We have shown that our model is much faster  than previous one-shot diffusion-based graph generation.  There are efforts on accelerating the sampling process of diffusion models \citep{de2021diffusion}. It remains an open problem to investigate how effective can such techniques accelerate diffusion-based graph generation without compromising generation performance. In addition, unlike our \ours model, these one-shot diffusion models need to maintain the full adjacency matrix in memory. This results in large memory footprints and can make them difficult to handle large graphs.

\section*{Acknowledgements}
We thank the anonymous reviewers for their helpful comments. This work was supported in part by the NSF (Expeditions CCF-1918770, CAREER IIS-2028586, IIS-2027862, IIS-1955883, IIS-2106961, IIS-2008334, CAREER IIS-2144338, PIPP CCF-2200269), CDC MInD program, faculty research award from Facebook and funds/computing resources from Georgia Tech.

\bibliography{icml2023_conference}
\bibliographystyle{icml2023}


\newpage
\appendix
\onecolumn
\section{Appendix}

\subsection{Visualization of generation ordering}
\label{sec:appendix:case}
\begin{figure}[h!]
\centering
  \includegraphics[width=\linewidth]{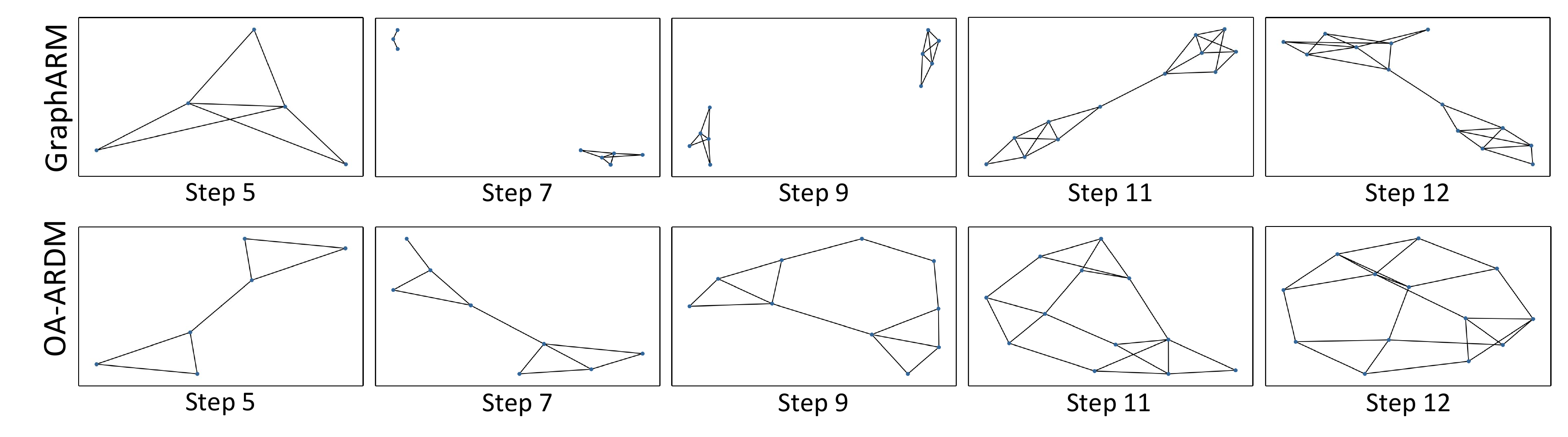}
\vspace{-3ex}
  \caption{The graph generative process of \ours and OA-ARDM for community generation. As we can see, \ours first generates one community and then adds another one, which show that \ours captures graph structural topology for generation. In contrast, OA-ARDM generates the graph with a random order. }
  \label{fig:case_study}
\end{figure}

Figure~\ref{fig:case_study} shows the graph generative process of GRAPHARM and OA-ARDM. As we can see,
GRAPHARM first generates one community and then moves to another; while OA-ARDM randomly
generates the graph and fails to capture the underlying graph distribution.

\subsection{Evaluation of Negative Log likelihood}
\label{sec:likelihood}

\begin{table}[h]
 \centering
  \setlength{\tabcolsep}{1mm}
\begin{tabular}{@{}lcc@{}}
\toprule
        & Community-small & Breast          \\ \midrule
GRAN    & $23.04$           & $247.18$          \\
OM    & $\bm{16.82}$  & $\bm{187.22}$ \\
EDP-GNN & N/A             & N/A             \\
GDSS    & N/A             & N/A             \\
\rowcolor[HTML]{EFEFEF}\ours    & $\bm{16.21}$  & $\bm{191.05}$ \\ \bottomrule
\end{tabular}
\caption{Test set negative log-likelihood (NLL) on Community-small and Breast datasets. N/A represents the model cannot provide the NLL. }
\label{table:nll}
\end{table}

We further evaluate the expected negative log-likelihood (NLL)  across node permutations on the test sets. For GRAN, we sample 1000 node permutations from the uniform distribution. For OM and \ours, we sample 1000 node permutations from the ordering network. Table~\ref{table:nll} shows the expected NLL on the test sets for community-small and Breast. As shown, \ours can achieve competitive results with OM and outperform GRAN by large margins. This is because both OM and \ours learn a data-dependent node ordering distribution; sampling from this distribution is more sample efficient than the uniform distribution. Though GDSS and EDP-GNN are also diffusion-based graph generative models, they cannot provide the likelihood. Note that GDSS involves a system of SDEs and we cannot directly use the equation from \citep{songsc2021} to compute the likelihood. To compute the likelihood, GDSS needs to derive the probability ODE flow that induces the same marginal probability of the system of SDEs. However, such a derivation is non-trivial and the GDSS paper did not provide it. This can be a drawback in  some density-based downstream tasks, \eg, outlier detection.

\subsection{Connection Between Autoregressive Diffusion and Absorbing Diffusion}
\label{app:arm}
While ARDM appears different from classic diffusion, it amounts to absorbing
diffusion with continuous time limit. Starting from state $\vx_0$, 
we can define a continuous-time absorbing process, where
each element
$\vx_t^{(i)}$ independently decays into an absorbing state with
continuous-time probabilities $\alpha(t)$. This stochastic process is equivalent as
using a finite set of $D$ random
transition times $\{\tau_i\}_{i=1}^{D}$ for recording the time where $\vx_t^{(i)}$ was
absorbed. It was shown by
\cite{hoogeboom2022autoregressive}
that 
modeling the reverse generation of this process does not need to be
conditioned on the precise values of the transition times
$\tau_i$. 
Hence, when training the reverse generative model, we only
need to model $\vx_{\tau_i}$ based on $\vx_{\tau_{i+1}}$ while ignoring
$\tau_i$.   This allows for  writing the variational lower bound
(VLB) of likelihood
as an expectation over an uniform ordering in an autoregressive form:
\begin{equation} \text{log} p(\vx_0) \geq \mathbb{E}_{\sigma \sim \mathcal{U}(S_d)} \sum_i^D \text{log}p(\vx_{\sigma(i)}|\vx_{\sigma(>i)}). \end{equation}

\subsection{Derivation of Eq.~3}
\label{sec:appendix:derivation}
The diffusion process assigns a unique ID to each node of $G_0$. Therefore, there is a one-to-one mapping between $G_{0:n}$ and $\sigma_{1:n}$, which leads to $p(\sigma_{1:n}|G_{0:n})=1$. Then, we have the following equations:
\begin{equation}
    p_{\theta}(G_{1:n},\sigma_{1:n})= p_{\theta}(G_{1:n}),
\end{equation}
\begin{equation}
q_{\phi}(G_{1:n}|G_0)=q_{\phi}(\sigma_{1:n}|G_0).
\end{equation}
Then, our VLB can be written as:
\begin{align}
  \log{p_{\theta}(G_{0})}  =&  \log{\left(\int p(G_{0:n})\frac{q_{\phi}(G_{1:n}|G_0)}{q_{\phi}(G_{1:n}|G_0)}dG_{1:n}\right)} \nonumber\\
                          \geq& \mathbb{E}_{q_{\phi}(G_{1:n}|G_0)}\left( \log{p(G_{1:n})} + \log{p(G_0|G_{1:n})} - \log{q_{\phi}(G_{1:n}|G_{0})} \right) \nonumber\\
                         =& \mathbb{E}_{q_{\phi}(\sigma_{1:n}|G_0)}\left( \log{p(G_{1:n},\sigma_{1:n})} + \log{p(G_0|G_{1:n})} - \log{q_{\phi}(G_{1:n}|G_{0})} \right) \nonumber
                          \\
                          =& \mathbb{E}_{q_{\phi}(\sigma_{1:n}|G_0)}( \log{p(G_{1:n-1}|\sigma_{1:n})p(\sigma_{1:n}|G_n)p(G_n)} + \log{p(G_0|G_{1:n})} - \log{q_{\phi}(\sigma_{1:n}|G_{0})} ) \nonumber\\
                              =& \mathbb{E}_{q_{\phi}(\sigma_{1:n}|G_0)} ( \log{p(G_{1:n-1}|\sigma_{1:n})} + \log{p(G_0|G_{1:n})} +\log{p(\sigma_{1:n}|G_n)} +\cancelto{0}{\log{p(G_n)}}
                                                     - \log{q_{
                              \phi}(\sigma_{1:n}|G_{0})} )
                            \nonumber \\
                                       =& \mathbb{E}_{q_{\phi}(\sigma_{1:n}|G_0)}\left( \log{p(G_{0:n-1}|\sigma_{1:n}, G_n)}+ \log{p(\sigma_{1:n}|G_n)}- \log{q_{\phi}(\sigma_{1:n}|G_{0})}
                            \right)\nonumber \\
                          =& \mathbb{E}_{q_{\phi}(\sigma_{1:n}|G_0)}
                            \sum_{t=0}^{n-1}\log{p_{\theta}(G_t|G_{t+1},\sigma_{t+1})}-\text{KL}(q_{\phi}(\sigma_{1:n}|G_0)|p_{\theta'}(\sigma_{1:n}|G_n)) \nonumber \\
                                      =& \mathbb{E}_{q_{\phi}(\sigma_{1:n}|G_0)}
                            \sum_{t=0}^{n-1}\log{p_{\theta}(G_t|G_{t+1})}-\text{KL}(q_{\phi}(\sigma_{1:n}|G_0)|p_{\theta'}(\sigma_{1:n}|G_n)).
\end{align}

 We have $\log p(G_n)=0$ because $G_n$ is a deterministic graph wherein all the nodes are masked. In the last line, we omit the $\sigma_t$ in $p_{\theta}(G_t|G_{t+1})$ because  $G_t$ together with $G_{t+1}$ contains the information that we want to predict the node $v_{\sigma_t}$ and its edges with previously denoised nodes.
 
\subsection{Differences from OM}
\label{sec:appendix:comparison}

The previous works \citep{chen2021order} first use the node ordering $\sigma$ to permute the adjacency matrix and then autoregressively grow the permuted adjacency matrix $A^{\sigma}$ row by row. However, the permuted adjacency matrix $A^{\sigma}$ does not contain the exact node ordering information $\sigma$ since multiple node orders can lead to the same adjacency matrix.  In OM, $p_\theta(A^{\sigma}_{1: n}, \sigma_{1: n})$ is factorized as: $$p_\theta\left(A^{\sigma}_{1: n}, \sigma_{1: n}\right) = p_\theta\left(A^{\sigma}_{1: n}\right) p_\theta\left(\sigma_{1: n} \mid A^{\sigma}_{1: n} \right).$$ As there are multiple $\sigma_{1:n}$ that lead to $A^{\sigma}_{1:n}$, OM inevitably needs to compute the graph automorphism in their VLB.

Why do we not need to compute the graph automorphism as \citep{chen2021order}? This is because, in our framework, the node ordering $\sigma_{1:n}$ is sampled in the forward diffusion process which essentially assigns a unique ID to each node of $G_0$. This creates a one-to-one mapping between $G_{0:n}$ and $\sigma_{1:n}$.
In the generation procedure, our starting point is the masked graph $G_n$ with $n$  nodes; our generator network sequentially denoises the node $v_{\sigma_t}$ in $G_n$ in the reverse order of the diffusion ordering during training. 

Different from OM, we factorize $p_\theta\left(G_{1: n}\right)$ (line 4 of Eq.~8) as:

$$ p\left(G_{1: n}, \sigma_{1: n}\right) = p\left(G_{1: n-1} \mid \sigma_{1: n}\right) p\left(\sigma_{1: n} \mid G_n\right) p\left(G_n\right).$$

This factorization leads to the following consequences / benefits:

- The challenge of modeling the node ordering variable is now pushed into the term $p\left(\sigma_{1: n} \mid G_n\right)$. This is another key reason that we do not have the graph automorphism term in our VLB, instead we have a KL term. Now how do we learn this conditional distribution? It turns out we do not need to! While the generation ordering network is required for non-graph data such as text to determine which token to unmask at test time, it is not needed for graph generation due to node permutation invariance. The first term in Eq.~\ref{eq:VLB} will encourage the denoising network $p_{\theta}(G_t|G_{t+1})$
to predict the node and edge types in the exact reverse ordering of the diffusion process, thus the denoising network itself can be a proxy of the generation ordering.
Due to permutation invariance, we can simply replace any masked node and its masked edges with the predicted node and edge types at each time step.  

- Still because of the factorization, our generator network $p_{\theta}(G_{0:n-1}|G_n,\sigma_{1:n})$ models the graph generation \emph{conditioned on} a fixed node ordering. In other words, given an oracle ordering of how the nodes should be generated (which is given by the diffusion ordering network $q$), we use the generator network to model how the graph sequences are generated. 

\subsection{Training algorithm of \ours}
\label{sec:appendix:alg}

We use a reinforcement learning (RL) procedure by sampling multiple diffusion
trajectories, thereby enabling training both the diffusion ordering network
$q_{\phi}(\sigma|G_0)$ and the denoising network $p_{\theta}(G_t|G_{t+1})$ using
gradient descent.

Specifically, at each training iteration, we explore the diffusion ordering
network by creating $M$ diffusion trajectories for each training graph $G^{(i)}_0$. Each trajectory is a sequence of graphs $\{G_t^{i,m}\}_{1\leq t\leq n}$ where the node decay ordering $\sigma^{i,m}$ is sampled from $q_{\phi}(\sigma|G_0^{i,m})$. For each trajectory, we sample $T$ time steps.
The denoising network
$p_{\theta}(G_t|G_{t+1})$ is then trained to minimize the negative VLB using stochastic gradient descent (SGD):
\begin{align}
\triangle \theta \leftarrow
  \frac{\eta_1}{M}\nabla \sum_{i\in\mathcal{B}_{\text{train}}}\sum_{ m,t}\sum_{k\in\sigma_{(\leq t)}}\frac{n_iw^{i,m}_{k}}{T}\log p_{\theta}(O_{v_{k}}^{\sigma_{(>t)}}|G^{i,m}_{t+1}), \nonumber
\end{align}
where $\mathcal{B}_{\rm train}$ is the a minibatch sampled from the training data and $w_k^{i,m}=q_{\phi}(\sigma_t^{i,m}=k|G_0^i, \sigma_{(<t)}^{i,m})$.

To evaluate the current
diffusion ordering network, we create $M$ trajectories for each validation graph and
compute the negative VLB of the
denoising network to obtain the corresponding rewards $R^{i,m}=-\sum_{t}\sum_{k\in \sigma_{(\leq t)}}\frac{n_i}{T}w_{k}^{i,m}\text{log} p_{\theta}(O_{v_{k}}^{\sigma_{(>t)}}|G^{i,m}_{t+1})$. Then, the diffusion ordering network can be updated with common RL optimization methods, \eg, the REINFORCE algorithm \citep{williams1992simple}:
\begin{equation}
  \triangle\phi \leftarrow \frac{\eta_2}{M} \sum_{i\in \mathcal{B}_{\text{val}}}\sum_{m} R^{i,m} \nabla \log q_{\phi}(\sigma|G_0^{i,m}).
  \label{eq:reinforce}
\end{equation}
 The detailed training procedure is summarized in Algorithm~\ref{alg:overall}.

\begin{algorithm}[h]
\caption{Training procedure of \ours}
\begin{algorithmic}
\REQUIRE Diffusion ordering network $q_{\phi}(\sigma|G_0)$, Denoising network $p_{\theta}(G_t|G_{t+1})$ 
\FOR {\# training iterations}
\STATE Sample a minibatch $\mathcal{B}$ from the training set 
\FOR {each $i\in \mathcal{B}_{\rm train}$}
\FOR {each $m\in [1,M]$}
\STATE $\sigma_1^{i,m},\cdots, \sigma_{n_i}^{i,m}\sim q_{\phi}(\sigma|G_0)$
\STATE Obtain corresponding diffusion trajectories $\{G_t^{i,m}\}_{1\leq t\leq n}$
\ENDFOR
\STATE Sample $T$ time steps from $\mathcal{U}_n$
\STATE    $\theta_{j} \leftarrow \theta_{j-1} -
  \frac{\eta_1}{M}\nabla \sum_{i\in\mathcal{B}_{\text{train}}}\sum_{ m,t}\sum_{k\in\sigma_{(\leq t)}}\frac{n_iw^{i,m}_{k}}{T}\log p_{\theta}(O_{v_{k}}^{\sigma_{(>t)}}|G^{i,m}_{t+1})$
\ENDFOR
\STATE Sample a minibatch from the validation set
\FOR {each $i\in \mathcal{B}_{\rm val}$}
\FOR {each $m\in [1,M]$}
\STATE $\sigma_1^{i,m},\cdots, \sigma_{n_i}^{i,m}\sim q_{\phi}(\sigma|G_0)$
\STATE Obtain corresponding diffusion trajectories $\{G_t^{i,m}\}_{1\leq t\leq n}$
\ENDFOR 
\STATE Sample $T$ time steps from $\mathcal{U}_n$
\STATE Compute the reward $R^{i,m}=-\sum_{t}\sum_{k\in \sigma_{(\leq t)}}\frac{n_i}{T}w_{k}^{i,m}\text{log} p_{\theta}(O_{v_{k}}^{\sigma_{(>t)}}|G^{i,m}_{t+1})$
\STATE $  \triangle\phi \leftarrow \frac{\eta_2}{M} \sum_{i\in \mathcal{B}_{\text{val}}}\sum_{m} R^{i,m} \nabla \log q_{\phi}(\sigma|G_0^{i,m})$.
\ENDFOR
\ENDFOR
\end{algorithmic}
\label{alg:overall}
\end{algorithm}

\subsection{Implementation Details on Generic Graphs}
\label{sec:appendix:implementation}

\textbf{Model optimization}: We use ADAM with $\beta_1=0.9$ and $\beta=0.999$ as the optimizer. The learning rate is set for $10^{-4}$ and $5\times 10^{-4}$ for the denoising network and diffusion ordering network respectively on all the datasets. We perform model selection based on the average MMD of the three metrics on the validation set.

\textbf{Network architecture}:
For fair comparison, our denoising network uses the same graph attention network architecture as the baseline GRAN which has 7 layers and hidden dimensions 128;
the diffusion ordering network uses the same graph attention network architecture as the baseline OA which is a vallina GAT \citep{velickovic2018graph} that has 3 layers with 6 attention heads and residual connections. The hidden dimension is set to 32. As OM is compatible with any existing autoregressive methods, we use the strongest autoregressive baseline GRAN as its backbone.

\textbf{Hyper-parameters}: We set the number of trajectories $M$ as 4 for all the datasets. Both \ours and GRAN use 20 as the number of Bernoulli mixtures. For GRAN, we use block size 1 and stride 1 to achieve the best generation performance. For GDSS, we choose the best signal-to-noise ratio (SNR) from $\{0.05, 0.1, 0.15, 0.2\}$ and scale coefficient from $\{0.1,0.2,0.3,0.4,0.5,0.6,0.7,0.8,0.9,1.0\}$
based on the average MMD of degree, clustering coefficient and orbit as in \cite{jo2022score}. For EDP-GNN, we use 6 noise levels $\{\sigma_{i}\}_{i=1}^L=[1.6,0.8,0.6,0.4,0.2,0.1]$ as suggested in the original work \citep{niu2020permutation}. For OM, we use 4 as the sample size for variational inference as suggested in the original work \citep{chen2021order}

\subsection{Baselines and Implementation Details on Molecule Datasets}
\label{sec:appendix:implementation:molecule}

\textbf{Baselines}:
We compare \ours with the following baselines: GraphAF~\citep{shi2019graphaf} and GraphDF~\citep{luo2021graphdf} are flow-based autoregressive models.
GraphEBM~\citep{liu2021graphebm} is a one-shot energy-based model. MoFlow~\citep{zang2020moflow} is a one-shot normalizing flow model. SPECTRE \citep{martinkus2022spectre} is a one-shot model based on GAN from a spectral perspective.
DiGress \citep{vignac2022digress},
EDP-GNN~\citep{niu2020permutation} and 
GDSS~\citep{jo2022score} are 
scored-based one-shot methods.

\textbf{Network architecture}:

Denosing network:

We first use a one linear layer MLP to project the node type and edge type into the continuous embedding space.
$$ \vh^0_{v_{i}}=\text{MLP}(v_{i}), \quad \vh_{e_{v_i,v_j}} = \text{MLP}(e_{v_i,v_j})$$
Then, the l-th round of message passing is implemented as:
$$\vm^l_{i,j}=f([\vh^l_{v_i}|\vh^l_{v_j}|\vh_{e_{v_i,v_j}}]), \quad a^l_{i,j}=\text{Sigmoid}(g([\vh^l_{v_i}|\vh^l_{v_j}|\vh_{e_{v_i,v_j}}])),$$
$$ \vh_{v_i}^{l+1}=\text{GRU}(\vh_{v_i}^l,\sum_{j\in \mathcal{N}(i)}a_{i,j}^l\vm^l_{i,j}),$$ 
where $[\cdot|\cdot]$ is the concatenate operation. Both $f$ and $g$ are 2-layer MLPs with ReLU nonlinearities and hidden dimension 256.

After $L=5$ rounds of message passing, we obtain the final node representation vectors $\vh^L_{v_i}$ for each node. With the average pooling operation, we can further obtain the graph level representation $\vh^L_{G}$.

Then the prediction of node $v_{\sigma_t}$ follows a multinomial distribution: 
$$p(v_{\sigma_t}|G_{t+1})=\text{Softmax}(\text{MLP}_{n}([\vh_G^L|\vh_{v_{\sigma_t}}]),$$
where we implement $\text{MLP}_{n}$ as a 2-layer MLPs with ReLU nonlinearities and hidden dimension 256.

The prediction of edges between $v_{\sigma_t}$ and all the previously denoised nodes $\{v_j\}_{j\in\sigma_{(>t)}}$ follows a mixture of Multinomial distribution:
$$p(e_{v_t,v_j}|G_{t+1})=\sum_{k=1}^K\alpha_k\text{Softmax}(\text{MLP}_{e_k}([\vh_G^L|\vh_{v_t}|\vh_{v_j}]),$$  
$$\alpha_{1}, \cdots, \alpha_{K} = \text{Softmax}(\sum_{j\in \sigma_{(>t)}}\text{MLP}_{\alpha}([\vh_G^L|\vh_{v_i}|\vh_{v_j}])).$$
 Both $\text{MLP}_{\alpha}$ and $\text{MLP}_{e_k}$ are 2-layer MLPs with ReLU nonlinearities and hidden dimension 256. We use $K=20$ in both experiments.

Diffusion ordering network: We use a 3-layer relational graph convolutional network with hidden dimension 256.

\textbf{Model optimization}: We use ADAM with $\beta_1=0.9$ and $\beta=0.999$ as the optimizer. The learning rate is set for $10^{-3}$ and $5\times 10^{-2}$ for the denoising network and diffusion ordering network respectively on both the datasets.

\subsection{Experiment Setup for Constrained Graph Generation}
\label{sec:appendix:constrained}
We use the Caveman dataset for the constrained generation experiment. We set
the constraint as that the maximum node degree is no larger than 6. To
generate graphs under this constraint with \ours, we add a degree checking
into the generation process. Specifically, when generating a new node and its
connecting edges, we first check whether the constraint will be violated on
previous nodes as new edges are added to them. If so, the corresponding edges
will be dropped; otherwise we proceed to check the constraint on the new
generated node and randomly remove the extra edges that exceed the limit.

\subsection{Dataset Statistics}
\label{sec:appendix:dataset}
We evaluate the performance of \ours on six diverse graph generation benchmarks, covering both synthetic and real-world graphs from different domains:
(1) Community-small \citep{you2018graphrnn}, containing 100 graphs randomly generated community graphs with $12\leq |V|\leq 20$;
(2) Caveman \citep{You}, containing 200 caveman graphs synthetically generated with $5\leq |V| \leq 10$;
(3) Cora, containing 200 sub-graphs with $9\leq |V| \leq 87$, extracted from Cora network~\citep{sen2008collective} using random walk;
(4) Breast, including 100 chemical graphs with $12\leq |V|\leq 18$, sampled from  \cite{gonzalez2011high};
(5) Enzymes, including 563 protein graphs with $10\leq |V|\leq 125$ from BRENDA database~\citep{schomburg2004brenda};
(6) Ego-small, containing 200 small sub-graphs with $4\leq |V|\leq 18$ sampled from Citeseer Network Dataset~\citep{sen2008collective}. For each dataset, we use $80\%$ of the graphs as training set and the rest $20\%$ as test sets. Following \citep{liao2019efficient}, we randomly select $25\%$ from the training data as the validation set.

Table~\ref{table:molecule:stats} provides the statistics of QM9 and ZINC250k datasets used in the molecule generation tasks.

\begin{table}[h]
\centering
\begin{tabular}{@{}ccccc@{}}
\toprule
Dataset  & Number of graphs & Number of nodes & Number of node types & Number of edge types \\ \midrule
QM9      & 133,885          & $1\leq |V|\leq 9$               & 4                    & 3                    \\
ZINC250k & 249,455          &  $6\leq |V|\leq 38$               & 9                    & 3                    \\ \bottomrule
\end{tabular}
\caption{Statistics of QM9 and ZINC250k datasets used in the molecule generation tasks.}
\label{table:molecule:stats}
\end{table}

\subsection{Verification of the Denoising ordering}
In Eq.~3, we do not learn a separate generation ordering but directly replace the masked node and edges with the predictions of the denoising network at test time. This is because graph is node permutation invariant and the first term in Eq.~3 encourages the denoising model to predict the node and edge types in the reverse of the diffusion ordering. To evaluate the generation ordering directly obtained from the denoising network, we compute the KL-divergence between the diffusion ordering and the denoising ordering at each time step and then sum together, i.e., $\text{Approximation error=}\sum_{t}\text{KL}\left(q(\sigma_t|G_0,\sigma_{(<t))}|p(\sigma_t|G_{t})\right)$. Specifically, at each time step $t$, we use \ours to first generate a graph $G_0$ and record the corresponding node generation ordering $\sigma$. Since we do not explicitly model distribution of the node index during the generation procedure, we determine the index of the generated node in $G_0$ by its connectivity to the unmasked nodes in $G_{t}$. Finally, we forward the generation network multiple times to obtain an empirical distribution for $p(\sigma_t|G_{t})$. Table~\ref{table:kl} provides the approximation error versus the training iterations. As we can see, with the training progresses, the approximation error indeed becomes smaller and approaches to zero.

\begin{table}[h]
\centering
\begin{tabular}{@{}lllllll@{}}
\toprule
Training iterations & 0     & 100   & 200   & 500   & 2000  & 3000  \\ \midrule
Appriximation error & 0.343 & 0.262 & 0.131 & 0.093 & 0.061 & 0.049 \\ \bottomrule
\end{tabular}
\caption{Approximation error between the generation ordering and the diffusion ordering on the community-small dataset. }
\label{table:kl}
\end{table}

\subsection{Visualization of Generated Samples}

\begin{figure}[h]
    \centering
    \includegraphics[width=0.9\linewidth]{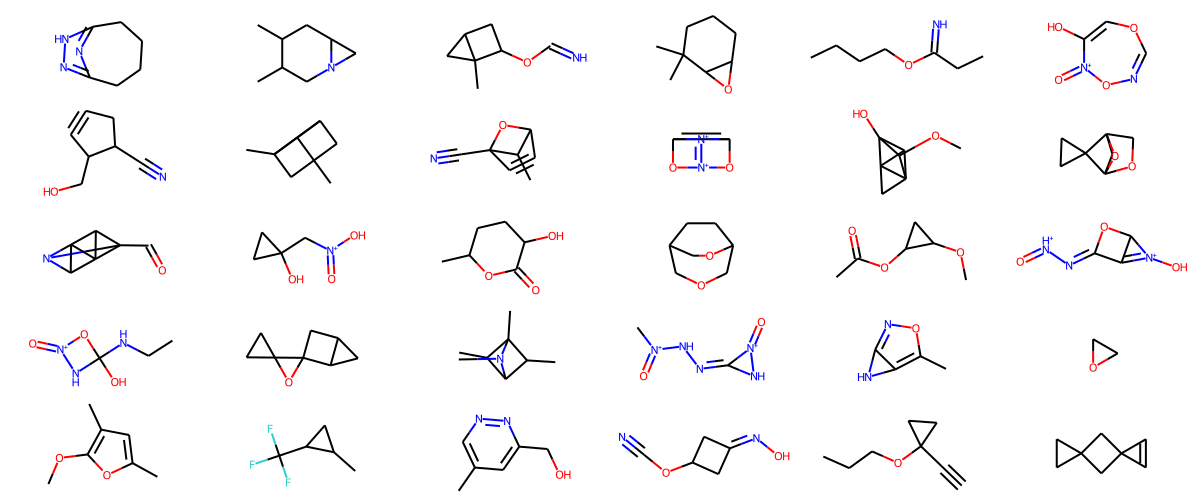}
    \caption{Visualization of generated samples of \ours on QM9 dataset}
    \label{fig:qm9}
\end{figure}

\begin{figure}[h]
    \centering
    \includegraphics[width=0.9\linewidth]{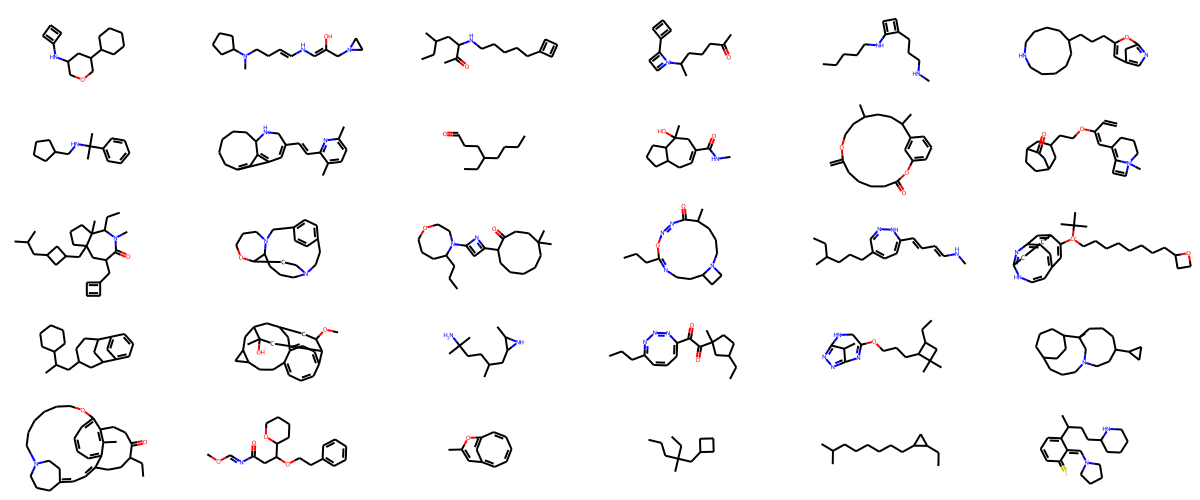}
    \caption{Visualization of generated samples of \ours on ZINC250K dataset}
    \label{fig:zinck}
\end{figure}

%%%%%%%%%%%%%%%%%%%%%%%%%%%%%%%%%%%%%%%%%%%%%%%%%%%%%%%%%%%%%%%%%%%%%%%%%%%%%%%
%%%%%%%%%%%%%%%%%%%%%%%%%%%%%%%%%%%%%%%%%%%%%%%%%%%%%%%%%%%%%%%%%%%%%%%%%%%%%%%

\end{document}